\documentclass[10pt,twocolumn,letterpaper]{article}

\usepackage{cvpr}              

\usepackage{amsmath}
\usepackage{amssymb}
\usepackage{booktabs}
\usepackage{graphicx}
\usepackage{ulem}
\usepackage{algorithm}
\usepackage{algorithmicx}
\usepackage{algpseudocode}
\usepackage{framed}

\usepackage{multirow}
\usepackage{float}

\floatname{algorithm}{Algorithm}

\makeatletter

\usepackage[pagebackref,breaklinks,colorlinks]{hyperref}

\usepackage[capitalize]{cleveref}
\crefname{section}{Sec.}{Secs.}
\Crefname{section}{Section}{Sections}
\Crefname{table}{Table}{Tables}
\crefname{table}{Tab.}{Tabs.}

\def\cvprPaperID{*****} 
\def\confName{CVPR}
\def\confYear{2022}

\begin{document}

\title{YOLOP: You Only Look Once for Panoptic Driving Perception}

\author{Dong Wu \quad Manwen Liao \quad Weitian Zhang \quad Xinggang Wang\\ 
\quad Xiang Bai \quad Wenqing Cheng \quad Wenyu Liu\\
School of EIC, Huazhong University of Science \& Technology\\
{\tt\small \{riserwu,mwliao,wtzhang,xgwang,xbai,chengwq,liuwy\}@hust.edu.cn}
}
\maketitle

\begin{abstract}
   A panoptic driving perception system is an essential part of autonomous driving. A high-precision and real-time perception system can assist the vehicle in making the reasonable decision while driving. We present a panoptic driving perception network (YOLOP) to perform traffic object detection, drivable area segmentation and lane detection simultaneously. It is composed of one encoder for feature extraction and three decoders to handle the specific tasks. Our model performs extremely well on the challenging BDD100K dataset, achieving state-of-the-art on all three tasks in terms of accuracy and speed. Besides, we verify the effectiveness of our multi-task learning model for joint training via ablative studies. To our best knowledge, this is the first work that can process these three visual perception tasks simultaneously in real-time on an embedded device Jetson TX2(23 FPS) and maintain excellent accuracy. To facilitate further research, the source codes and pre-trained models are released at \url{https://github.com/hustvl/YOLOP}
\end{abstract}

\section{Introduction}
\label{sec:intro}

Recently, extensive research on autonomous driving has revealed the importance of panoptic driving perception system. It plays a significant role in autonomous driving as it can extract visual information from the images taken by the camera and assist the decision system to control the actions of the vehicle. In order to restrict the maneuver of vehicles, the visual perception system should be able to understand the scene and then provide the decision system with information including: locations of the obstacles, judgements of whether the road is drivable, the position of the lanes etc. Object detection is usually involved in the panoptic driving perception system to help the vehicles avoid obstacles and follow traffic rules. Drivable area segmentation and lane detection are also needed as they are crucial for planning the driving route of the vehicle.

\begin{figure}
\centering
\begin{subfigure}{\linewidth}
  \includegraphics[width=\linewidth]{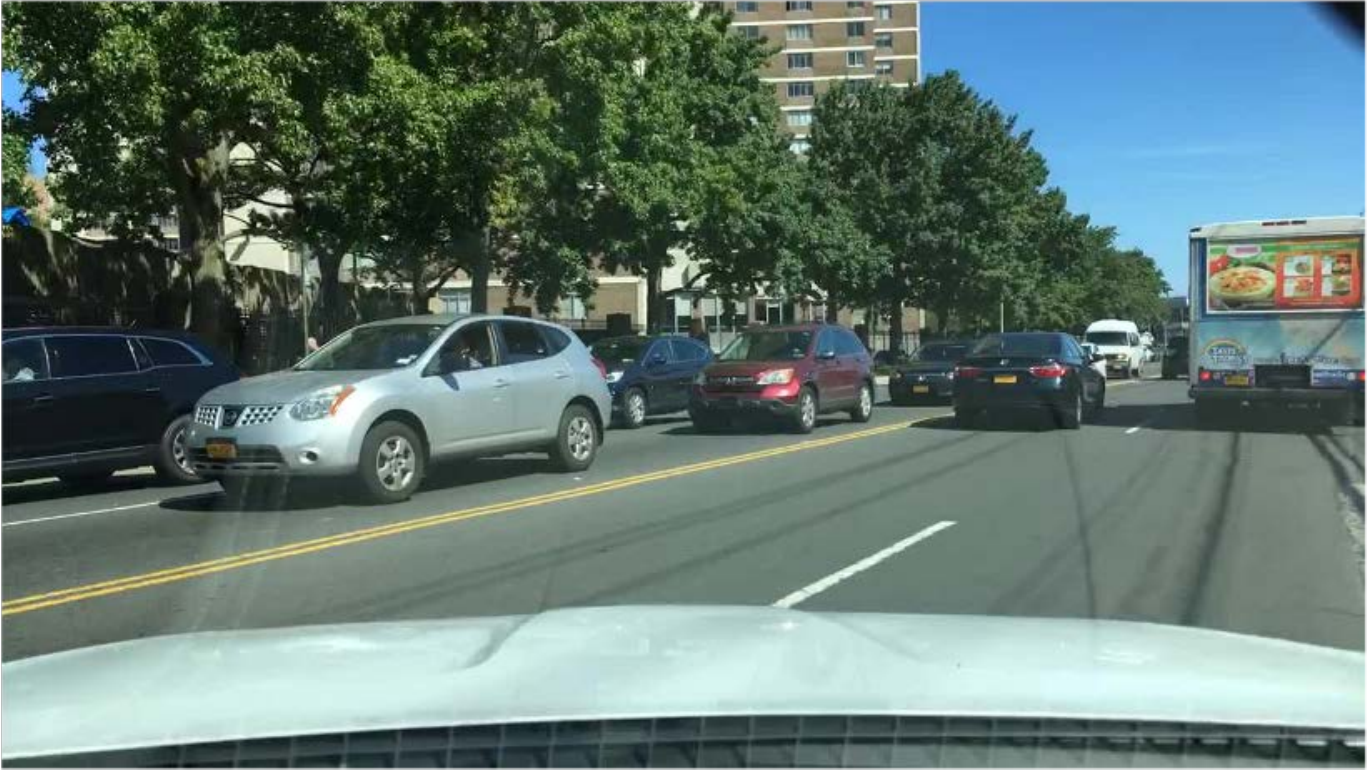}
  \caption{Input}
\end{subfigure}

\bigskip
\begin{subfigure}{\linewidth}
  \includegraphics[width=\linewidth]{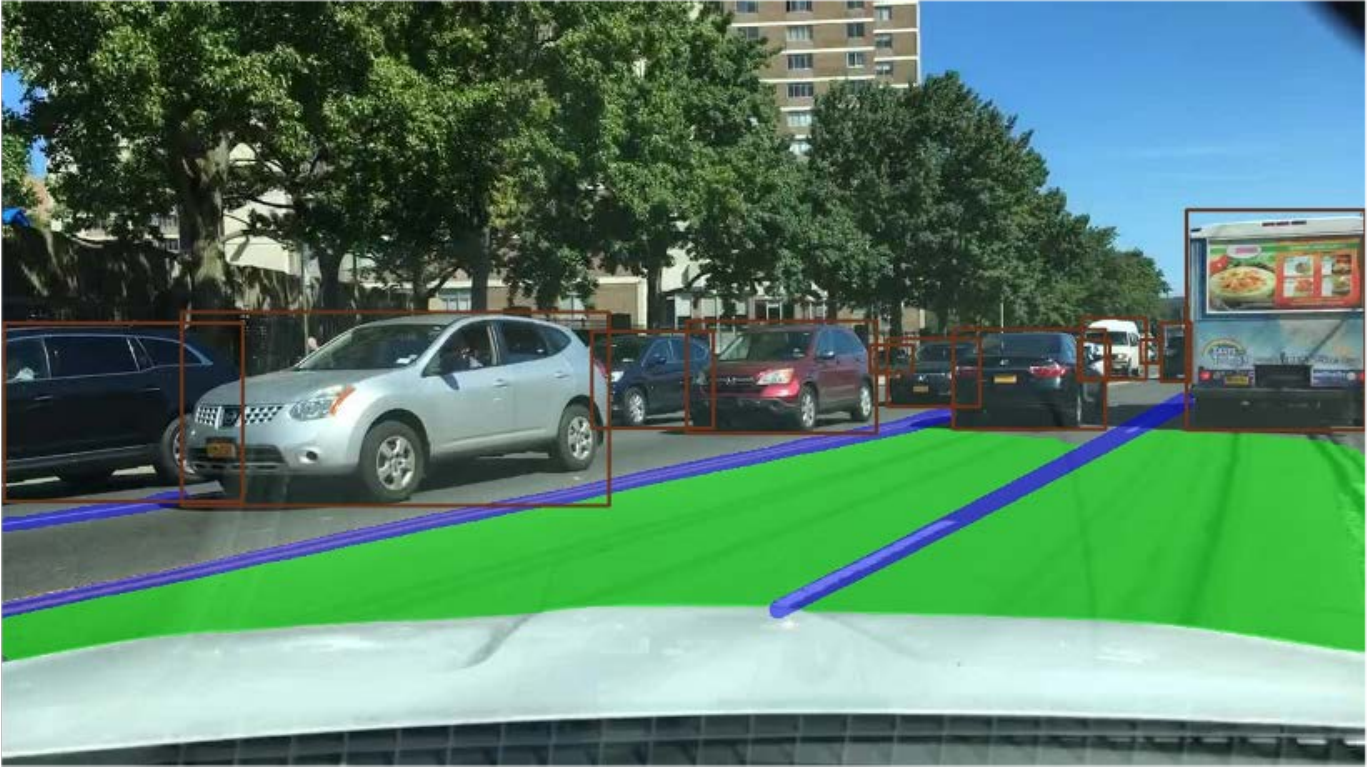}
  \caption{Output}
\end{subfigure}
\caption{ The input and output of our model. The purpose of our model is to process traffic objects detection, drivable area segmentation and lane detection simultaneously in one input image. In (b), the brown bounding boxes indicate traffic objects, the green areas are the drivable areas, and the blue lines represent the lane line.}
\label{fig:total}
\end{figure}

For such a panoptic driving perception system, high-precision and real-time are two most critical requirements, which are related to whether the autonomous vehicle can make accurate and timely decision to ensure safety. However, for practical autonomous driving system, especially the ADAS, the computational resources are often marginal and limited. Therefore, it is very challenging to take both requirements into account in real-world scenarios.

Many methods handle these tasks separately. For instance, Faster R-CNN \cite{faster-rcnn} and YOLOv4 \cite{yolov4} deal with object detection; ENet \cite{enet} and PSPNet \cite{pspnet} are proposed to perform semantic segmentation. SCNN \cite{scnn} and SAD-ENet \cite{sad-enet} are used for detecting lanes. Despite the excellent performance these methods achieve, processing these tasks one after another takes longer time than tackling them all at once. When deploying the panoptic driving perception system on embedded devices commonly used in the self-driving car, limited computational resources and latency should be taken into account. In addition, different tasks in traffic scenes understanding often have much related information. As shown in Figure \ref{fig:total}, lanes are often the boundary of drivable area, and drivable area usually closely surrounds the traffic objects. A multi-task network is more suitable in this situation as (1) it can accelerate the image analysis process by handling multiple tasks simultaneously 
rather than sequentially. (2) it can share information among multiple tasks as multi-task network often shares the same feature extraction backbone. Therefore, it is of essence to explore multi-task approaches in autonomous driving.

In order to solve the multi-task problem for panoptic driving perception, i.e., traffic object detection, drivable area segmentation and lane detection, while obtaining high precision and fast speed, we design a simple and efficient network architecture. We use a lightweight CNN \cite{cspdarknet} as the encoder to extract features from the image. Then these feature maps are fed to three decoders to complete their respective tasks. Our detection decoder is based on the current best-performing single-stage detection network \cite{yolov4} for two main reasons: (1) The single-stage detection network is faster than the two-stage detection network. (2) The grid-based prediction mechanism of the single-stage detector is more related to the other two semantic segmentation tasks, while instance segmentation is usually combined with the region-based detector \cite{mask-rcnn}. And we verify the two viewpoints in the  experiments section. The feature map output by the encoder incorporates semantic features of different levels and scales, and our segmentation branch can use these feature maps to complete pixel-wise semantic prediction excellently.

In addition to the end-to-end training strategy, we attempt some alternating optimization paradigms which train our model step-by-step. On the one hand, we can put unrelated tasks in different training steps to prevent inter-limitation. On the other hand, the task trained first can guide other tasks. So this kind of paradigm sometimes works well though cumbersome. However, experiments show that it is unnecessary for our model as the one trained end to end can perform well enough. Our panoptic driving perception system reaches 41 FPS on a single NVIDIA TITAN XP and 23 FPS on Jetson TX2; meanwhile, it achieves state-of-the-art on the three tasks of the BDD100K dataset \cite{bdd100k}.

In summary, our main contributions are: 
(1) We put forward an efficient multi-task network that can jointly handle three crucial tasks in autonomous driving: object detection, drivable area segmentation and lane detection to save computational costs and reduce inference time. Our work is the first to reach real-time on embedded devices while maintaining state-of-the-art level performance on the BDD100K dataset.
(2) We design the ablative experiments to verify the effectiveness of our multi-tasking scheme. It is proved that the three tasks can be learned jointly without tedious alternating optimization.
(3) We design the ablative experiments to prove that the grid-based prediction mechanism of detection task is more related to that of semantic segmentation task, which is believed to provide reference for other relevant multi-task learning research works.

\section{Related Work}
In this section, we review solutions to the above three tasks respectively, and then introduce some related multi-task learning work. We only concentrate on solutions based on deep learning.

\subsection{Traffic Object Detection}
In recent years, with the rapid development of deep learning, many prominent object detection algorithms have emerged. Current mainstream object detection algorithms can be divided into two-stage methods and one-stage methods.

Two-stage methods complete the detection task in two steps. First, regional proposals are obtained, and then features in the regional proposals are used to locate and classify the objects. The generation of regional proposals has gone through several stages of development \cite{rcnn,fast-rcnn,faster-rcnn,r-fcn}. 

The SSD-series \cite{ssd} and YOLO-series algorithms are milestones among one-stage methods. This kind of algorithm performs bounding box regression and object classification simultaneously. YOLO \cite{yolo} divides the picture into S$\times$S grids instead of extracting regional proposals with the RPN network, which significantly accelerates the detection speed.
YOLO9000 \cite{yolov2} introduces the anchor mechanism to improve the recall of detection. YOLOv3 \cite{yolov3} uses the feature pyramid network structure to achieve multi-scale detection. YOLOv4 \cite{yolov4} further improves the detection performance by refining the network structure, activation function, loss function and applying abundant data augmentation.

\subsection{Drivable Area Segmentation}
Due to the rapid development of deep learning, a number of CNN-based methods have made great success in semantic segmentation area, and they can be applied in drivable area segmentation task to provide pixel-level results. FCN \cite{fcn} firstly introduces fully convolutional network to semantic segmentation. Despite the skip-connection refinement, its performance is still limited by low-resolution output. PSPNet \cite{pspnet} comes up with the pyramid pooling module to extract features in various scales to enhance its performance. Besides accuracy, speed is also a key element in evaluating this task. In order to achieve real-time inference speed, ENet \cite{enet} reduces size of the feature maps. Recently, multitask learning is introduced to deal with this task, EdgeNet \cite{edgenet} combine edge detection with drivable area segmentation task to obtain more accurate segmentation results without compromising the inference speed.

\subsection{Lane Detection}
In lane detection, there are lots of innovative researches based on deep learning. \cite{lanenet} constructs a dual-branch network to perform semantic segmentation and pixel embedding on images. It further clusters the dual-branch features to achieve lane instance segmentation. SCNN \cite{scnn} proposes slice-by-slice convolution, which enables the message to pass between pixels across rows and columns in a layer, but this convolution is very time-consuming. Enet-SAD \cite{sad-enet} uses self attention distillation method, which enables low-level feature maps to learn from high-level feature maps. This method improves the performance of the model while keeping the model lightweight.

\subsection{Multi-task Approaches}
The goal of multi-task learning is to learn better representations through shared information among multiple tasks. Especially, a CNN-based multitask learning method can also achieve convolutional sharing of the network structure. Mask R-CNN \cite{mask-rcnn} extends Faster R-CNN by adding a branch for predicting object mask, which combines instance segmentation and object detection tasks effectively, and these two tasks can promote each other's performance. LSNet \cite{LSNet} summarizes object detection, instance segmentation and pose estimation as location-sensitive visual recognition and uses a unified solution to handle these tasks. With a shared encoder and three independent decoders, MultiNet \cite{multinet} completes the three scene perception tasks of scene classification, object detection and segmentation of the driving area simultaneously. DLT-Net \cite{dlt-net} inherits the encoder-decoder structure, and contributively constructs context tensors between sub-task decoders to share designate information among tasks. \cite{zhang2018geometric} puts forward mutually interlinked sub-structures between lane area segmentation and lane boundary detection. Meanwhile, it proposes a novel loss function to constrain the lane line to the outer contour of the lane area so that they're going to overlap geometrically. However, this prior assumption also limits its application as it only works well on scenarios where the lane line tightly wraps the lane area. What's more, the training paradigm of multitask model is also worth thinking about. \cite{kang2011learning} states that the joint training is appropriate and beneficial only when all those tasks are indeed related; otherwise, it is necessary to adopt alternating optimization. So Faster R-CNN \cite{faster-rcnn} adopts a pragmatic 4-step training algorithm to learn shared features. This paradigm sometimes may be helpful, but mostly it is tedious.

\section{Methodology}

\begin{figure*}
\begin{center}
\includegraphics[width=0.95\linewidth]{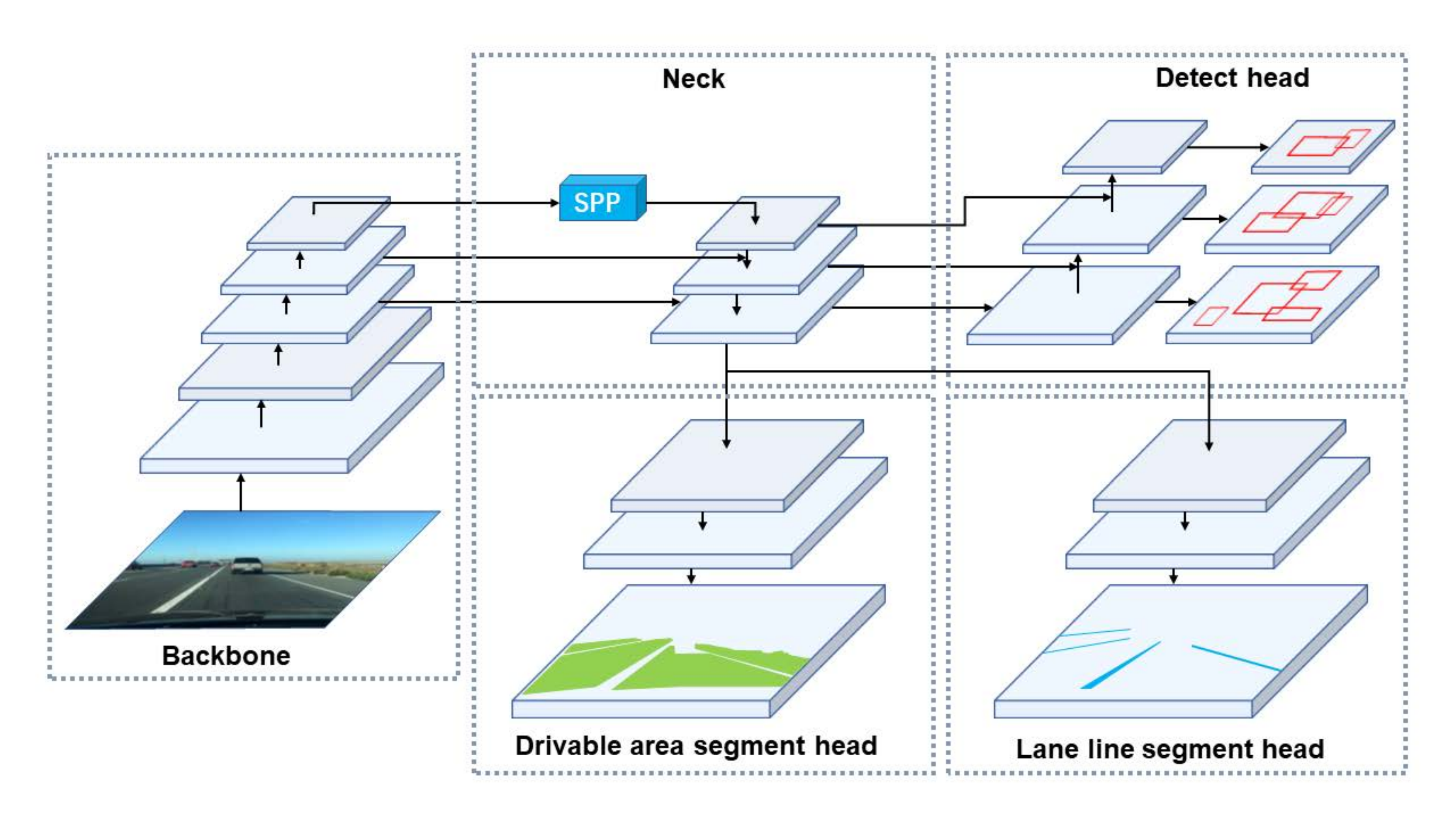}
\end{center}
   \caption{The architecture of YOLOP. YOLOP shares one encoder and combines three decoders to solve different tasks. The encoder consists of a backbone and a neck.}
\label{fig:network}
\end{figure*}

We put forward a simple and efficient feed-forward network that can accomplish traffic object detection, drivable area segmentation and lane detection tasks altogether. As shown in Figure \ref{fig:network}, our panoptic driving perception  single-shot network, termed as YOLOP, contains one shared encoder and three subsequent decoders to solve specific tasks. There are no complex and redundant shared blocks between different decoders, which reduces computational consumption and allows our network to be easily trained end-to-end.

\subsection{Encoder}
Our network shares one encoder, which is composed of a backbone network and a neck network.

\subsubsection{Backbone}
The backbone network is used to extract the features of the input image. Usually, some classic image classification networks serve as the backbone. Due to the excellent performance of YOLOv4 \cite{yolov4} on object detection, we choose CSPDarknet \cite{cspdarknet} as the backbone, which solves the problem of gradient duplication during optimization \cite{cspnet}. It supports feature propagation and feature reuse which reduces the amount of parameters and calculations. Therefore, it is conducive to ensuring the real-time performance of the network.

\subsubsection{Neck}
The neck is used to fuse the features generated by the backbone. Our neck is mainly composed of Spatial Pyramid Pooling (SPP) module \cite{sppnet} and Feature Pyramid Network (FPN) module \cite{fpn}. SPP generates and fuses features of different scales, and FPN fuses features at different semantic levels, making the generated features contain multiple scales and multiple semantic level information. We adopt the method of concatenation to fuse features in our work.

\subsection{Decoders}
The three heads in our network are specific decoders for the three tasks.

\subsubsection{Detect Head}
Similar to YOLOv4, we adopt an anchor-based multi-scale detection scheme. Firstly, we use a structure called Path Aggregation Network (PAN), a bottom-up feature pyramid network \cite{pannet}. FPN transfers semantic features top-down, and PAN transfers positioning features bottom-up. We combine them to obtain a better feature fusion effect, and then directly use the multi-scale fusion feature maps in the PAN for detection. Then, each grid of the multi-scale feature map will be assigned three prior anchors with different aspect ratios, and the detection head will predict the offset of position and the scaling of the height and width, as well as the corresponding probability of each category and the confidence of the prediction.

\subsubsection{Drivable Area Segment Head \& Lane Line Segment Head}
Drivable area segment head and Lane line Segment head adopt the same network structure. We feed the bottom layer of FPN to the segmentation branch, with the size of $(W/8,H/8,256)$. Our segmentation branch is very simple. After three upsampling processes, we restore the output feature map to the size of $(W,H,2)$, which represents the probability of each pixel in the input image for the drivable area/lane line and the background.
Because of the shared SPP in the neck network, we do not add an extra SPP module to segment branches like others usually do \cite{pspnet}, which brings no improvement to the performance of our network. Additionally, we use the Nearest Interpolation method in our upsampling layer to reduce computation cost instead of deconvolution. As a result, not only do our segment decoders gain high precision output, but also be very fast during inference.

\subsection{Loss Function}
Since there are three decoders in our network, our multi-task loss contains three parts. As for the detection loss $\mathcal L_{det}$, it is a weighted sum of classification loss, object loss and bounding box loss as in equation \ref{eq:det_loss}.
\begin{equation} \label{eq:det_loss}
    \mathcal L_{det} = \alpha_1 \mathcal L_{class} + \alpha_2 \mathcal L_{obj} + \alpha_3 \mathcal L_{box},
\end{equation}
where $\mathcal L_{class}$ and $\mathcal L_{obj}$ are focal loss \cite{focalloss}, which is utilized to reduce the loss of well-classified examples, thus forces the network to focus on the hard ones. $\mathcal L_{class}$ is used for penalizing classification and $\mathcal L_{obj}$ for the confidence of one prediction. $\mathcal L_{box}$ is  $\mathcal L_{CIoU}$ \cite{d/ciouloss}, which takes distance, overlap rate, the similarity of scale and aspect ratio between the predicted box and ground truth into consideration.

Both of the loss of drivable area segmentation $\mathcal L_{da-seg}$ and lane line segmentation $\mathcal L_{ll-seg}$ contain Cross Entropy Loss with Logits $\mathcal L_{ce}$, which aims to minimize the classification errors between pixels of network outputs and the targets. It is worth mentioning that IoU loss: $\mathcal L_{IoU}=1-\frac{TP}{TP+FP+FN}$ is added to $\mathcal L_{ll-seg}$ as it is especially efficient for the prediction of the sparse category of lane lines. $\mathcal L_{da}$ and $\mathcal L_{ll-seg}$  are defined as equation (\ref{eq:da_loss}), (\ref{eq:lane_loss}) respectively.
\begin{equation} \label{eq:da_loss}
    \mathcal L_{da-seg} = \mathcal L_{ce},
\end{equation}
\begin{equation} \label{eq:lane_loss}
    \mathcal L_{ll-seg} = \mathcal L_{ce} + \mathcal L_{IoU}.
\end{equation}

In conclusion, our final loss is a weighted sum of the three parts all together as in equation (\ref{eq:loss}).
\begin{equation} \label{eq:loss}
    \mathcal L_{all} = \gamma_1 \mathcal L_{det} + \gamma_2 \mathcal L_{da-seg} + \gamma_3 \mathcal L_{ll-seg},
\end{equation}
where $\alpha_1, \alpha_2, \alpha_3, \gamma_1, \gamma_2, \gamma_3$  can be tuned to balance all parts of the total loss.

\subsection{Training Paradigm}

We attempt different paradigms to train our model. The simplest one is training end to end, and then three tasks can be learned jointly. This training paradigm is useful when all tasks are indeed related. In addition, some alternating optimization algorithms also have been tried, which train our model step by step. In each step, the model can focus on one or multiple related tasks regardless of those unrelated. Even if not all tasks are related, our model can still learn adequately on each task with this paradigm. And Algorithm \ref{algorithm:train} illustrates the process of one step-by-step training method.

\begin{algorithm}  
    \caption{One step-by-step Training Method. First, we only train Encoder and Detect head. Then we freeze the Encoder and Detect head as well as train two Segmentation heads. Finally, the entire network is trained jointly for all three tasks.} 
    \label{algorithm:train}
    \begin{algorithmic}[1] 
        \Require Target neural network $\mathcal{F}$ with parameter group:
        \Statex \quad \, $\Theta=\{\theta_{enc}, \theta_{det}, \theta_{seg}\}$;
        \Statex \quad \, Training set: $\mathcal T$;
        \Statex \quad \, Threshold for convergence: $thr$;
        \Statex \quad \, Loss function: $\mathcal L_{all}$
        \Ensure Well-trained network: $\mathcal{F}(\mathbf x;\Theta)$
        \Procedure{Train}{$\mathcal{F}$, $\mathcal T$}
            \Repeat
                \State Sample a mini-batch $(\mathbf x_s, \mathbf y_s)$ from training set $\mathcal T$.
                \State $\ell \leftarrow \mathcal L_{all}(\mathcal F(\mathbf x_s;\Theta), \mathbf y_s)$
                \State $\Theta \leftarrow \arg \min_{\Theta} \ell$
            \Until $\ell < thr$
        \EndProcedure
        \State $\Theta \leftarrow \Theta \setminus \{\theta_{seg}\}$ // Freeze parameters of two Segmentation heads.
        \State \Call{Train}{$\mathcal{F}$, $\mathcal T$}
        \State $\Theta \leftarrow \Theta \cup \{\theta_{seg}\} \setminus \{\theta_{det}, \theta_{enc}\}$ // Freeze parameters of Encoder and Detect head and activate parameters of two Segmentation heads.
        \State \Call{Train}{$\mathcal{F}$, $\mathcal T$}
        \State $\Theta \leftarrow \Theta \cup \{\theta_{det}, \theta_{enc}\}$ // Activate all parameters of the neural network.
        \State \Call{Train}{$\mathcal{F}$, $\mathcal T$}
        \State \Return {Trained network $\mathcal{F}(\mathbf{x};\Theta)$}
    \end{algorithmic}
\end{algorithm}


\section{Experiments}

\subsection{Setting}

\subsubsection{Dataset Setting}
The BDD100K dataset \cite{bdd100k} supports the research of multi-task learning in the field of autonomous driving. With 100k frames of pictures and annotations of 10 tasks, it is the largest driving video dataset. As the dataset has the diversity of geography, environment, and weather, the algorithm trained on the BDD100k dataset is robust enough to migrate to a new environment. Therefore, we choose the BDD100k dataset to train and evaluate our network. The BDD100K dataset has three parts, training set with 70K images, validation set with 10K images, and test set with 20K images. Since the label of the test set is not public, we evaluate our network on the validation set.

\subsubsection{Implementation Details}

In order to enhance the performance of our model, we empirically adopt some practical techniques and methods of data augmentation.

With the purpose of enabling our detector to get more prior knowledge of the objects in the traffic scene, we use the k-means clustering algorithm to obtain prior anchors from all detection frames of the dataset. We use Adam as the optimizer to train our model and the initial learning rate, ${\beta_1}$, and ${\beta_2}$ are set to be 0.001, 0.937, and 0.999 respectively. Warm-up and cosine annealing are used to adjust the learning rate during the training, which aim at leading the model to converge faster and better \cite{warmupandcos}.

We use data augmentation to increase the variability of images so as to make our model robust in different environments. Photometric distortions and geometric distortions are taken into consideration in our training scheme. For photometric distortions, we adjust the hue, saturation and value of images. We use random rotating, scaling, translating, shearing, and left-right flipping to process images to handle geometric distortions.

\subsubsection{Experimental Setting}

We select some excellent multi-task networks and networks that focus on a single task to compare with our network. Both MultiNet and DLT-Net handle multiple panoptic driving perception tasks, and they have achieved great performance in object detection and drivable area segmentation tasks on the BDD100k dataset. Faster-RCNN is an outstanding representative of the two-stage object detection network. YOLOv5 is the single-stage network that achieves state-of-the-art performance on the COCO dataset. PSPNet achieves splendid performance on semantic segmentation task with its superior ability to aggregate global information. We retrain the above networks on the BDD100k dataset and compare them with our network on object detection and drivable area segmentation tasks. Since there is no suitable existing multi-task network that processes lane detection task on the BDD100K dataset, we compare our network with Enet \cite{enet}, SCNN and Enet-SAD, three advanced lane detection networks. Besides, the performance of the joint training paradigm is compared with alternating training paradigms of many kinds. Moreover, we compare the accuracy and speed of our multi-task model trained to handle multiple tasks with the one trained to perform a specific task. Furthermore, we compare the performance of semantic segmentation task combined with single-stage detection task and two-stage detection task. Following \cite{sad-enet}, we resize images in BDD100k dataset from 1280$\times$720$\times$3 to 640$\times$384$\times$3. All control experiments follow the same experimental settings and evaluation metrics, and all experiments are run on NVIDIA GTX TITAN XP.

\subsection{Result}

In this section, we just simply train our model end to end and then compare it with other representative models on all three tasks.

\subsubsection{Traffic Object Detection Result}
\begin{figure*}
\centering
\begin{subfigure}{\linewidth}
  \includegraphics[width=\linewidth]{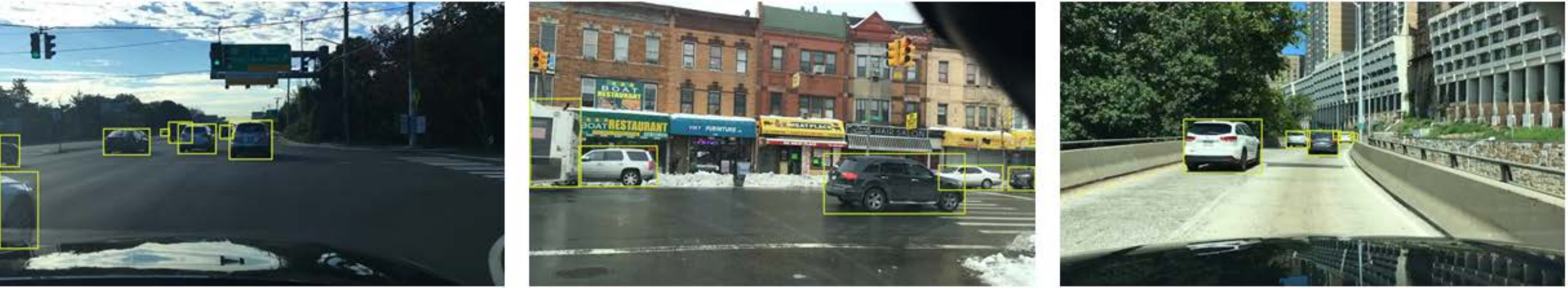}
  \caption{Day-time result}
\end{subfigure}

\begin{subfigure}{\linewidth}
  \includegraphics[width=\linewidth]{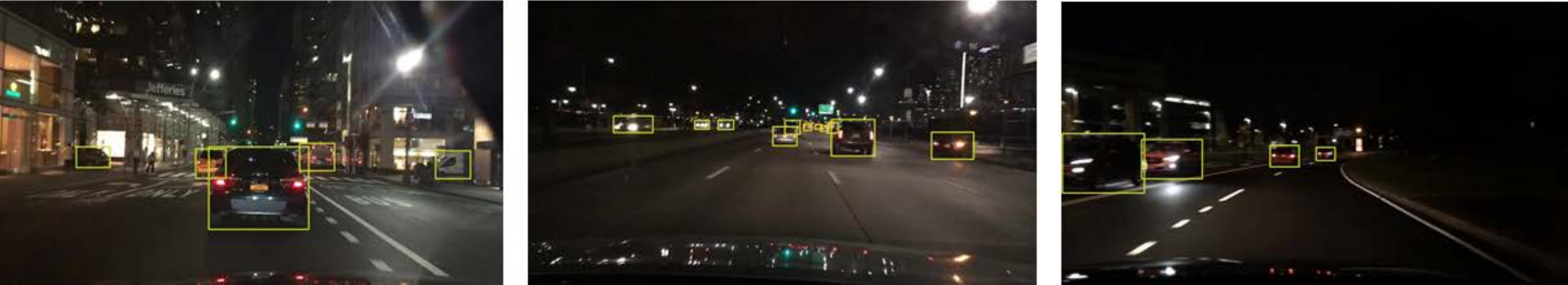}
  \caption{Night-time result}
\end{subfigure}
\caption{Visualization of the traffic objects detection results of YOLOP. Top Row: Traffic objects detection results in day-time scenes. Bottom row: Traffic objects detection results in night scenes.}
\label{fig:detect}
\end{figure*}

Visualization of the traffic objects detection is shown in Figure \ref{fig:detect}. Since the Multinet and DLT-Net can only detect vehicles, we only consider the vehicle detection results of five models on the BDD100K dataset. As shown in Table \ref{table:det_result}, we use Recall and mAP50 as the evaluation metric of detection accuracy. Our model exceeds Faster R-CNN, MultiNet, and DLT-Net in detection accuracy, and is comparable to YOLOv5s that actually uses more tricks than ours. Moreover, our model can infer in real time. YOLOv5s is faster than ours because it does not have the lane line segment head and drivable area segment head.

\begin{figure*}
\centering
\begin{subfigure}{\linewidth}
  \includegraphics[width=\linewidth]{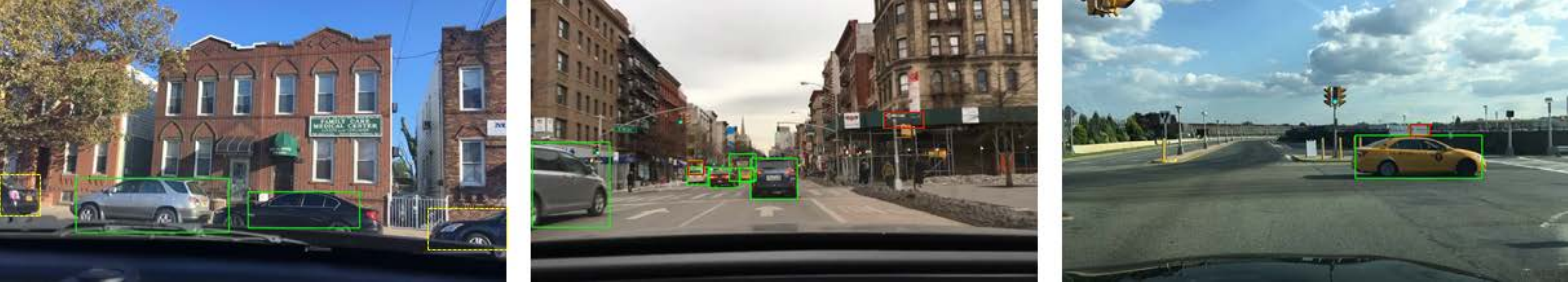}
  \caption{Results of Faster R-CNN}
\end{subfigure}

\begin{subfigure}{\linewidth}
  \includegraphics[width=\linewidth]{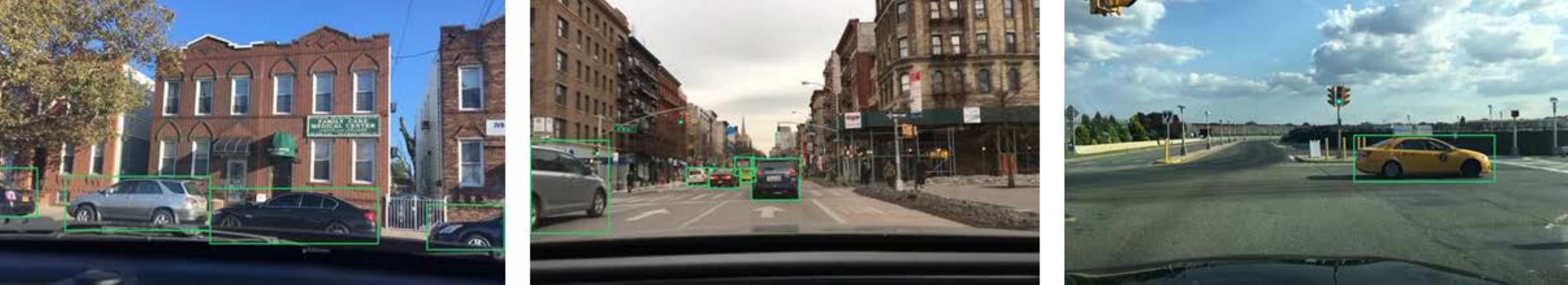}
  \caption{Results of YOLOP}
\end{subfigure}
\caption{Comparison between the traffic objects detection results of Faster R-CNN and YOLOP. Top Row: Traffic objects detection results of Faster R-CNN. Bottom row: Traffic objects detection results of YOLOP. The green bounding boxes are the detected correct vehicles. The yellow dotted bounding boxes are the false negative. The red  bounding boxes indicate the false positive.}
\label{fig:detect_vs}
\end{figure*}

Figure \ref{fig:detect_vs} shows the qualitative comparison between Faster R-CNN and YOLOP. Due to the information share of multi-task, the prediction results of YOLOP are more reasonable. For example, YOLOP will not misidentify the objects far from the road as vehicle. Moreover, the examples of false negative are much less and the bounding boxes are more accurate.

\begin{table}
\begin{center}
\setlength{\tabcolsep}{1.5mm}{
\begin{tabular}{cccc}
\toprule
Network & Recall(\%) & mAP50(\%) & Speed(fps)\\
\midrule
MultiNet & 81.3 & 60.2 & 8.6\\
DLT-Net & 89.4 & 68.4 & 9.3\\
Faster R-CNN & 81.2 & 64.9 & 8.8\\
YOLOv5s & 86.8 & 77.2 & 82\\
YOLOP (ours) & 89.2 & 76.5 & 41\\
\bottomrule
\end{tabular}
}
\end{center}
\caption{Traffic Object Detection Results: comparing the proposed YOLOP with state-of-the-art detectors.}
\label{table:det_result}
\end{table}

\subsubsection{Drivable Area Segmentation Result}
\begin{figure*}
\centering
\begin{subfigure}{\linewidth}
  \includegraphics[width=\linewidth]{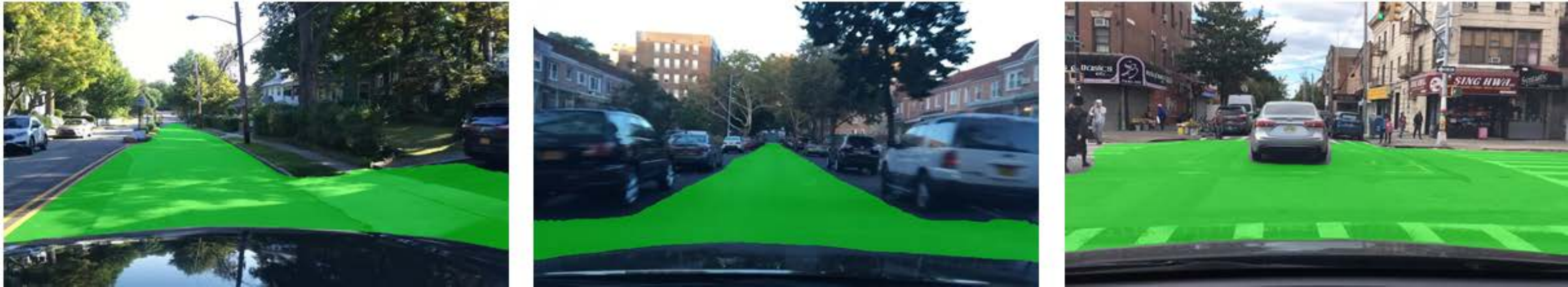}
  \caption{Day-time result}
\end{subfigure}

\begin{subfigure}{\linewidth}
  \includegraphics[width=\linewidth]{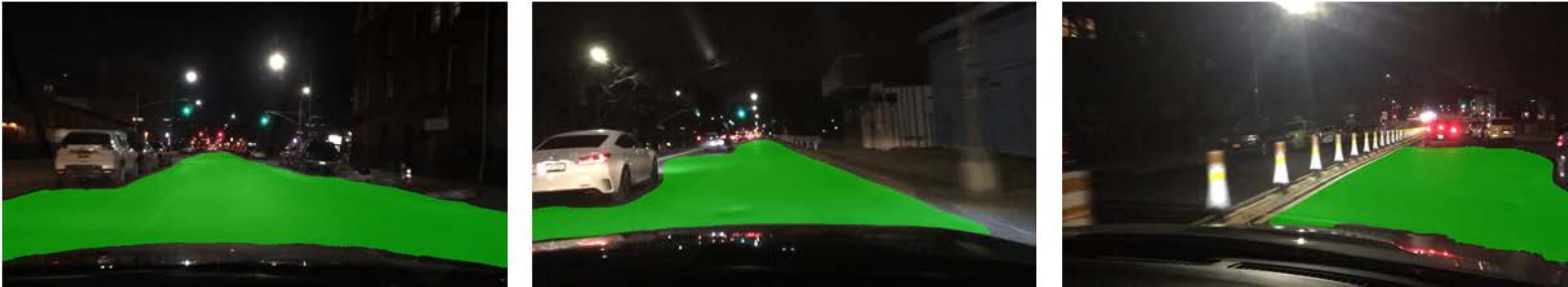}
  \caption{Night-time result}
\end{subfigure}
   \caption{Visualization of the drivable area segmentation results of YOLOP. Top Row: Drivable area segmentation results in day-time scenes. Bottom row: Drivable area segmentation results in night scenes.}
\label{fig:da}
\end{figure*}

Visualization results of the drivable area segmentation can be seen in Figure \ref{fig:da}. In this paper, both ``area/drivable" and ``area/alternative" classes in BDD100K dataset are categorized as "Drivable area" without distinction. Our model only needs to distinguish the drivable area and the background in the image. mIoU is used to evaluate the segmentation performance of different models. The results are shown in Table \ref{table:da_result}. It can be seen that our model outperforms MultiNet, DLT-Net and PSPNet by 19.9\%, 20.2\%, and 1.9\%, respectively. Furthermore, our inference speed is 4 to 5 times faster than theirs.

\begin{figure*}
\centering
\begin{subfigure}{\linewidth}
  \includegraphics[width=\linewidth]{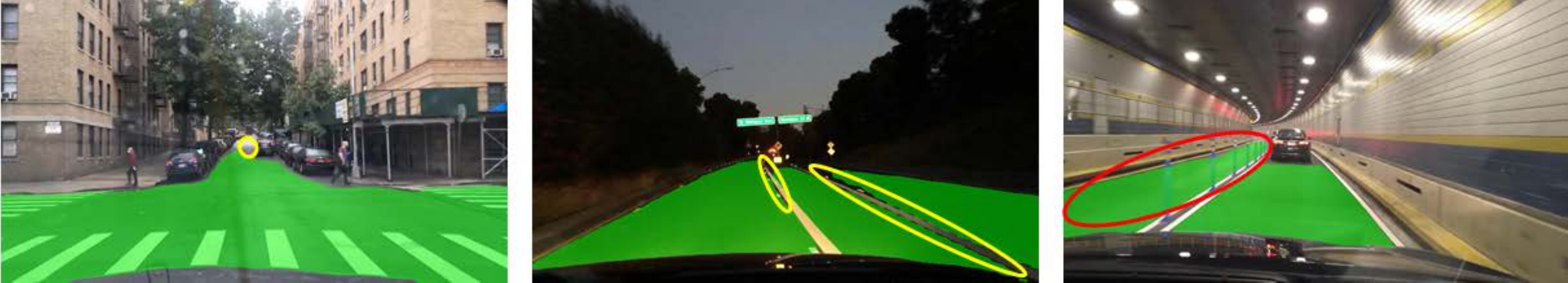}
  \caption{Results of PSPNet}
\end{subfigure}

\begin{subfigure}{\linewidth}
  \includegraphics[width=\linewidth]{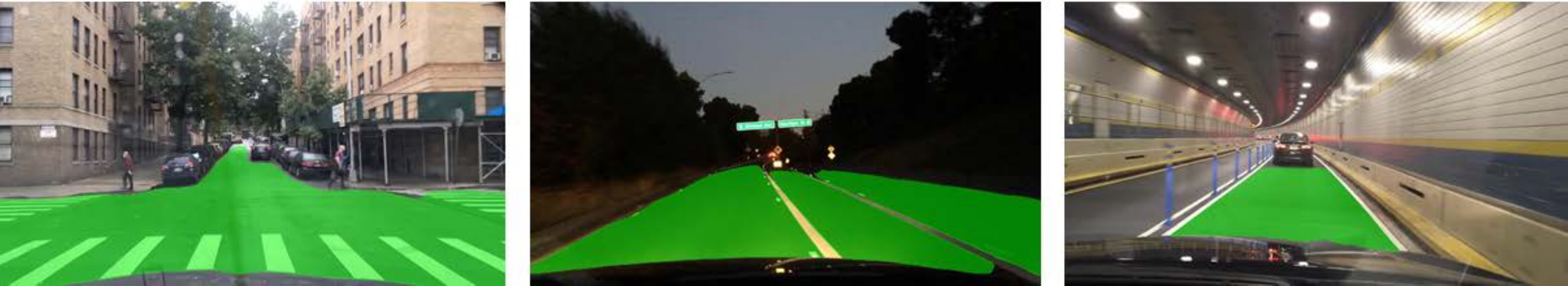}
  \caption{Results of YOLOP}
\end{subfigure}
\caption{Comparison between the drivable area segmentation results of PSPNet and YOLOP. Top Row: Drivable area segmentation results of PSPNet. Bottom row: Drivable area segmentation results of YOLOP. The yellow ellipses are the false negative. The red  ellipses indicate the false positive.}
\label{fig:da_vs}
\end{figure*}

The comparison between results of PSPNet and YOLOP is showed in Figure \ref{fig:da_vs}. Both PSPNet and YOLOP have perfomed well in this task. But YOLOP is significantly better at segmenting the edge areas that next to vehicles or lane lines. We think it's mainly because that both two other tasks provide the edge information for this task. Meanwhile, YOLOP makes fewer stupid mistakes, such as misjudging the opposite lane area as drivable area.

\begin{table}
\begin{center}
\setlength{\tabcolsep}{5mm}{
\begin{tabular}{ccc}
\toprule
Network & mIoU(\%) & Speed(fps)\\
\midrule
MultiNet & 71.6 & 8.6\\
DLT-Net & 71.3 & 9.3\\
PSPNet & 89.6 & 11.1\\
YOLOP (ours) & 91.5 & 41\\
\bottomrule
\end{tabular}
}
\end{center}
\caption{Drivable Area Segmentation Results: Comparing the proposed YOLOP with state-of-the-art drivable area segmentation or semantic segmentation methods.}
\label{table:da_result}
\end{table}

\subsubsection{Lane Detection Result}
\begin{figure*}
\centering
\begin{subfigure}{\linewidth}
  \includegraphics[width=\linewidth]{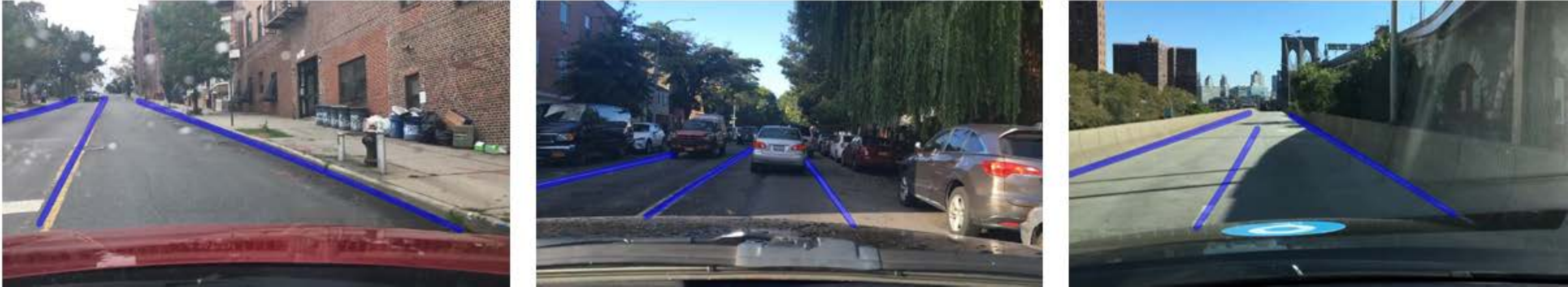}
  \caption{Day-time result}
\end{subfigure}

\begin{subfigure}{\linewidth}
  \includegraphics[width=\linewidth]{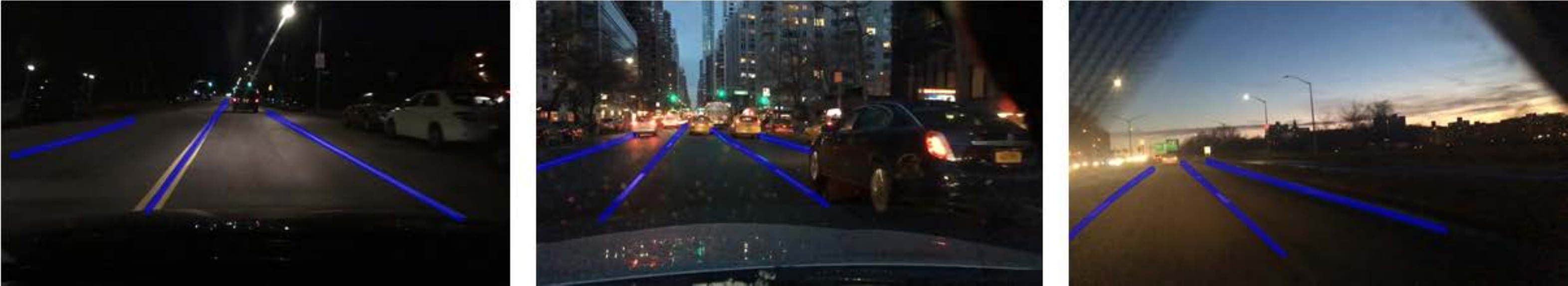}
  \caption{Night-time result}
\end{subfigure}
   \caption{Visualization of the lane detection results of YOLOP. Top Row: Lane detection results in day-time scenes. Bottom row: Lane detection results in night scenes.}
\label{fig:ll}
\end{figure*}

The visualization results of lane detection can be seen in Figure \ref{fig:ll}. The lane lines in BDD100K dataset are labeled with two lines, so it is very tricky to directly use the annotation. The experimental settings follow the \cite{sad-enet} in order to compare expediently. First of all, we calculate the center lines based on the two-line annotations. Then we draw the lane line of the training with width set to 8 pixels while keeping the lane line width of the test set as 2 pixels. We use pixel accuracy and IoU of lanes as evaluation metrics. As shown in the Table \ref{table:ll_result}, the performance of our model dramatically exceeds the other three models.

\begin{figure*}
\centering
\begin{subfigure}{\linewidth}
  \includegraphics[width=\linewidth]{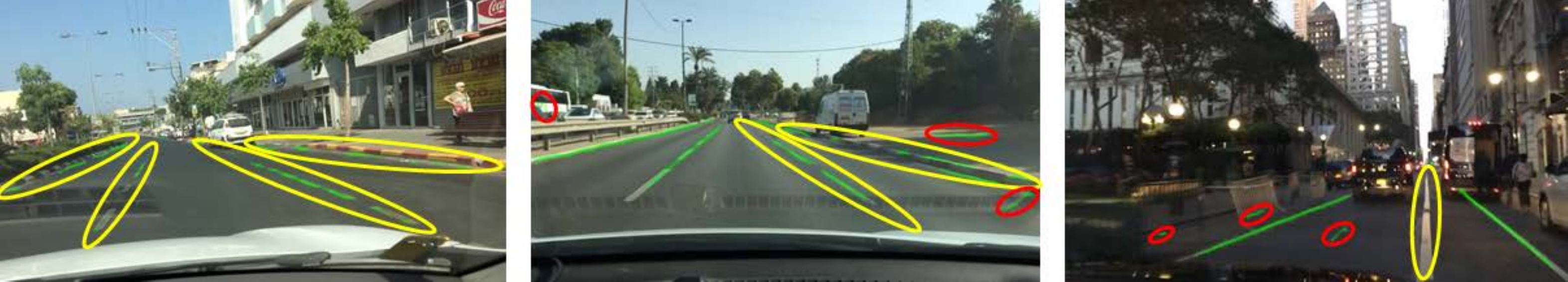}
  \caption{Results of ENet-SAD}
\end{subfigure}

\begin{subfigure}{\linewidth}
  \includegraphics[width=\linewidth]{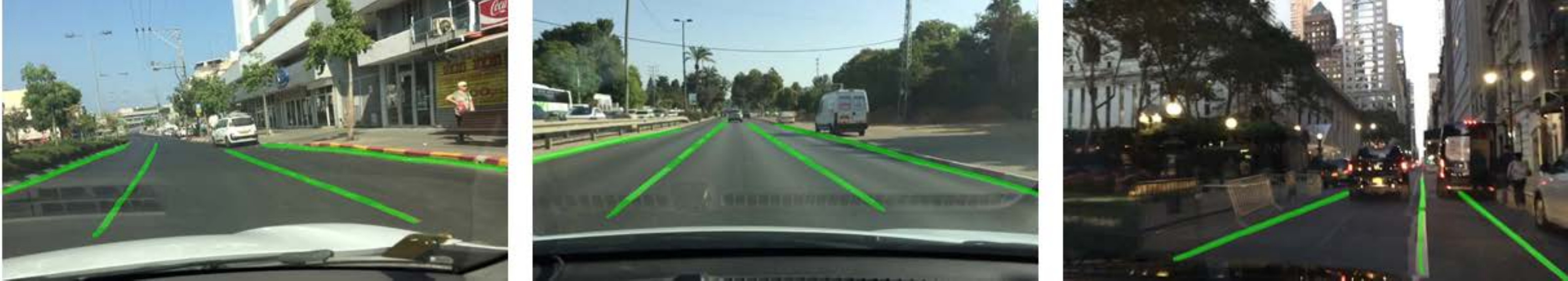}
  \caption{Results of YOLOP}
\end{subfigure}
\caption{Comparison between the lane detection results of ENet-SAD and YOLOP. Top Row: Lane detection results of ENet-SAD. Bottom row: Lane detection results of YOLOP. The yellow ellipses are the false negative. The red  ellipses indicate the false positive.}
\label{fig:ll_vs}
\end{figure*}

Figure \ref{fig:ll_vs} shows the comparison of Lane line detection results of ENet-SAD and YOLOP. The segmentation results of YOLOP is more accurate and continuous than ENet-SAD obviously. With the imformation shared by the other two tasks, YOLOP will not mistake some areas where some vehicles are located or driveable as lane lines, but Enet-SAD always does.

\begin{table}
\begin{center}
\setlength{\tabcolsep}{1.5mm}{
\begin{tabular}{cccc}
\toprule
Network & Accuracy(\%) & IoU(\%) & Speed(fps)\\
\midrule
ENet & 34.12 & 14.64 & 100\\
SCNN & 35.79 & 15.84 & 19.8\\
ENet-SAD & 36.56 & 16.02 & 50.6\\
YOLOP (ours) & 70.50 & 26.20 & 41\\
\bottomrule
\end{tabular}
}
\end{center}
\caption{Lane Detection Results: comparing the proposed YOLOP with state-of-the-art lane detection methods.}
\label{table:ll_result}
\end{table}

\subsection{Ablation Studies}
We designed the following three ablation experiments to further illustrate the effectiveness of our scheme. All the evaluation metrics in this section are consistent with above.

\subsubsection{End-to-end v.s. Step-by-step}
In Table \ref{table:EVS}, we compare the performance of joint training paradigm with alternating training paradigms of many kinds \footnote{E, D, S and W refer to Encoder, Detect head, two Segment heads and whole network. So the Algorithm \ref{algorithm:train} can be marked as ED-S-W, and the same for others.}. Obviously, our model has performed very well enough through end-to-end training, so there is no need to perform alternating optimization. However, it is interesting that the paradigm training detection task firstly seems to perform better. We think it is mainly because our model is closer to a complete detection model and the model is harder to converge when performing detection tasks. What's more, the paradigm consist of three steps slightly outperforms that with two steps. Similar alternating training can be run for more steps, but we have observed negligible improvements.

\begin{table*}[bh]
\begin{center}
\setlength{\tabcolsep}{7.5mm}{
\begin{tabular}{cccccc}
\toprule
Training method  & Recall(\%) & AP(\%) & mIoU(\%) & Accuracy(\%) & IoU(\%)\\
\midrule
ES-W & 87.0 & 75.3 & 90.4 & 66.8 & 26.2\\
ED-W & 87.3 & 76.0 & 91.6 & 71,2 & 26.1\\
ES-D-W & 87.0 & 75.1 & 91.7 & 68.6 & 27.0\\
ED-S-W & 87.5 & 76.1 & 91.6 & 68.0 & 26.8\\
End-to-end & 89.2 & 76.5 & 91.5 & 70.5 & 26.2\\
\bottomrule
\end{tabular}
}
\end{center}
\caption{Panoptic driving perception results: the end-to-end scheme v.s. different step-by-step schemes.}
\label{table:EVS}
\end{table*}

\subsubsection{Multi-task v.s. Single task}
To verify the effectiveness of our multi-task learning scheme, we compare the performance of the multi-task scheme and single task scheme. On the one hand, we train our model to perform 3 tasks simultaneously. On the other hand, we train our model to perform traffic object detection, drivable area segmentation, and lane line segmentation tasks separately. Table \ref{table:MVS} shows the comparison of the performance of these two schemes on each specific task. It can be seen that our model adopts the multi-task scheme to achieve performance is close to that of focusing on a single task. More importantly, the multitask model can save a lot of time compared to executing each task individually.

\subsubsection{Region-based v.s. Grid-based}
To verify the viewpoint that the grid-based prediction mechanism is more related to the two semantic segmentation tasks than the region-based prediction mechanism. We extend Faster R-CNN by adding two semantic segment heads to perform three tasks in parallel as our model did, and we call such a new model R-CNNP. We train both YOLOP and R-CNNP to (i) perform detection task and two segmentation tasks separately and (ii) three tasks simultaneously. In both two experiments above, the two segmentation tasks are trained jointly as there is no need to consider the interaction between them. All the experimental settings are the same, and the results are shown in Table \ref{table:GVR}. In the R-CNNP framework, the performance of multi-task training is much worse compared with training the detection task and semantic segmentation tasks separately. Obviously, the combination of two kinds of tasks conflicts in R-CNNP framework. But there is no such problem in our YOLOP framework, the performance of multi-task training is equal to that of focusing only on detection or semantic segmentation task. Thus we hold the opinion that this is due to the detection head of YOLOP, like two other semantic segmentation heads, directly perform global classification or regression tasks on the whole feature map output by Encoder, so they are similar and related in terms of prediction mechanism. Nevertheless, the detection head of R-CNNP needs to select region proposals first, and then perform prediction on the feature maps of each individual proposals, which is quite different from the global prediction mechanism of semantic segmentation. In addition, R-CNNP is far behind YOLOP in terms of inference speed. Therefore, our framework is a better choice for joint training detection and segmentation tasks.

\begin{table*}
\begin{center}
\setlength{\tabcolsep}{4.5mm}{
\begin{tabular}{ccccccc}
\toprule
Training method  & Recall(\%) & AP(\%) & mIoU(\%) & Accuracy(\%) & IoU(\%) & Speed(ms/frame)\\
\midrule
Det(only) & 88.2 & 76.9 & - & - & - & 15.7\\
Da-Seg(only) & - & - & 92.0 & - & - & 14.8\\
Ll-Seg(only) & - & - & - & 79.6 & 27.9 & 14.8\\
Multitask & 89.2 & 76.5 & 91.5 & 70.5 & 26.2 & 24.4\\
\bottomrule
\end{tabular}
}
\end{center}
\caption{Panoptic driving perception results: multi-task learning v.s. single task learning.}
\label{table:MVS}
\end{table*}

\begin{table*}
\renewcommand\arraystretch{1.1}
\begin{center}
\setlength{\tabcolsep}{4mm}{
\begin{tabular}{cccccccc}
\toprule
\multicolumn{2}{c}{Training method}  & Recall(\%) & AP(\%) & mIoU(\%) & Accuracy(\%) & IoU(\%) & Speed(ms/frame)\\
\midrule
\multirow{3}{*}{\rotatebox{90}{R-CNNP}} & Det(only) & 79.0 & 67.3 & - & - & - & -\\
~ & Seg(only) & - & - & 90.2 & 59.5 & 24.0 & -\\
~ & Multitask & 77.2(\textcolor{magenta}{-1.8}) & 62.6(\textcolor{magenta}{-4.7}) & 86.8(\textcolor{magenta}{-3.4}) & 49.8(\textcolor{magenta}{-9.7}) & 21.5(\textcolor{magenta}{-2.5}) & \textcolor{magenta}{103.3}\\
\midrule
\multirow{3}{*}{\rotatebox{90}{YOLOP}} & Det(only) & 88.2 & 76.9 & - & - & - & -\\
~ & Seg(only) & - & - & 91.6 & 69.9 & 26.5 & -\\
~ & Multitask & 89.2(\textcolor{green}{+1.0}) & 76.5(\textcolor{magenta}{-0.4}) & 91.5(\textcolor{magenta}{-0.1}) & 70.5(\textcolor{green}{+0.6}) & 26.2(\textcolor{magenta}{-0.3}) & \textcolor{green}{24.4}\\
\bottomrule
\end{tabular}
}
\end{center}
\caption{Panoptic driving perception results: Grid-based v.s. Region-based.}
\label{table:GVR}
\end{table*}

\section{Conclusion}
In this paper, we put forward a brand-new, simple and efficient network, which can simultaneously handle three driving perception tasks of object detection, drivable area segmentation and lane detection and can be trained end-to-end. Our model performs exceptionally well on the challenging BDD100k dataset, achieving or greatly exceeding state-of-the-art level on all three tasks. And it is the first to realize real-time reasoning on embedded device Jetson TX2, which ensures that our network can be used in real-world scenarios. Moreover, we have verified that the grid-based prediction mechanism is more related to that of semantic segmentation task. which may be of certain reference significance to similar multi-task learning research works.

Currently, although our multi-task network can be trained end-to-end without compromising the performance of each other, we hope to improve the performance of those tasks with more appropriate paradigm for multitask learning. Furthermore, our model is limited in three tasks, more tasks related with autonomous driving perception system such as depth estimation can be added in our future frame work to make the whole system more complete and practical.

{\small
\bibliographystyle{ieee_fullname}
\bibliography{Main}
}

\end{document}